\documentclass[10pt,twocolumn,letterpaper]{article}

\usepackage[pagenumbers]{cvpr} 
\usepackage{url}

\usepackage{paralist} 
\usepackage{multirow} 
\usepackage{soul} 

\usepackage{colortbl}

\newcommand{\csection}[1]{
\section{#1}
}

\newcommand{\csubsection}[1]{
\subsection{#1}
}

\definecolor{RubineRed}{RGB}{209,0,86}
\newcommand{\xhdr}[1]{\vspace{2.0pt}\noindent\textbf{#1}}

\definecolor{cvprblue}{rgb}{0.21,0.49,0.74}
\usepackage[pagebackref,breaklinks,colorlinks,allcolors=cvprblue]{hyperref}

\def\paperID{10871} 
\def\confName{CVPR}
\def\confYear{2025}

\title{Do Visual Imaginations Improve Vision-and-Language Navigation Agents?}

\author{
Akhil Perincherry\textsuperscript{1}, Jacob Krantz, Stefan Lee\textsuperscript{1}\\
\textsuperscript{1}Oregon State University\\
{\tt\small perincha@oregonstate.edu, jkrantz989@gmail.com, leestef@oregonstate.edu} 
}

\begin{document}
\maketitle

\begin{abstract}

Vision-and-Language Navigation (VLN) agents are tasked with navigating an unseen environment using natural language instructions. In this work, we study if visual representations of sub-goals implied by the instructions can serve as navigational cues and lead to increased navigation performance. To synthesize these visual representations or ``imaginations", we leverage a text-to-image diffusion model on landmark references contained in segmented instructions. These imaginations are provided to VLN agents as an added modality to act as landmark cues and an auxiliary loss is added to explicitly encourage relating these with their corresponding referring expressions. Our findings reveal an increase in success rate (SR) of $\sim$1 point and up to $\sim$0.5 points in success scaled by inverse path length (SPL) across agents. These results suggest that the proposed approach reinforces visual understanding compared to relying on language instructions alone. Code and data for our work can be found at \url{https://www.akhilperincherry.com/VLN-Imagine-website/}.

\end{abstract}

\csection{Introduction}
\label{sec:intro}

\begin{figure}[t]
    \centering
    \vspace{25pt}
    \includegraphics[width=\linewidth]{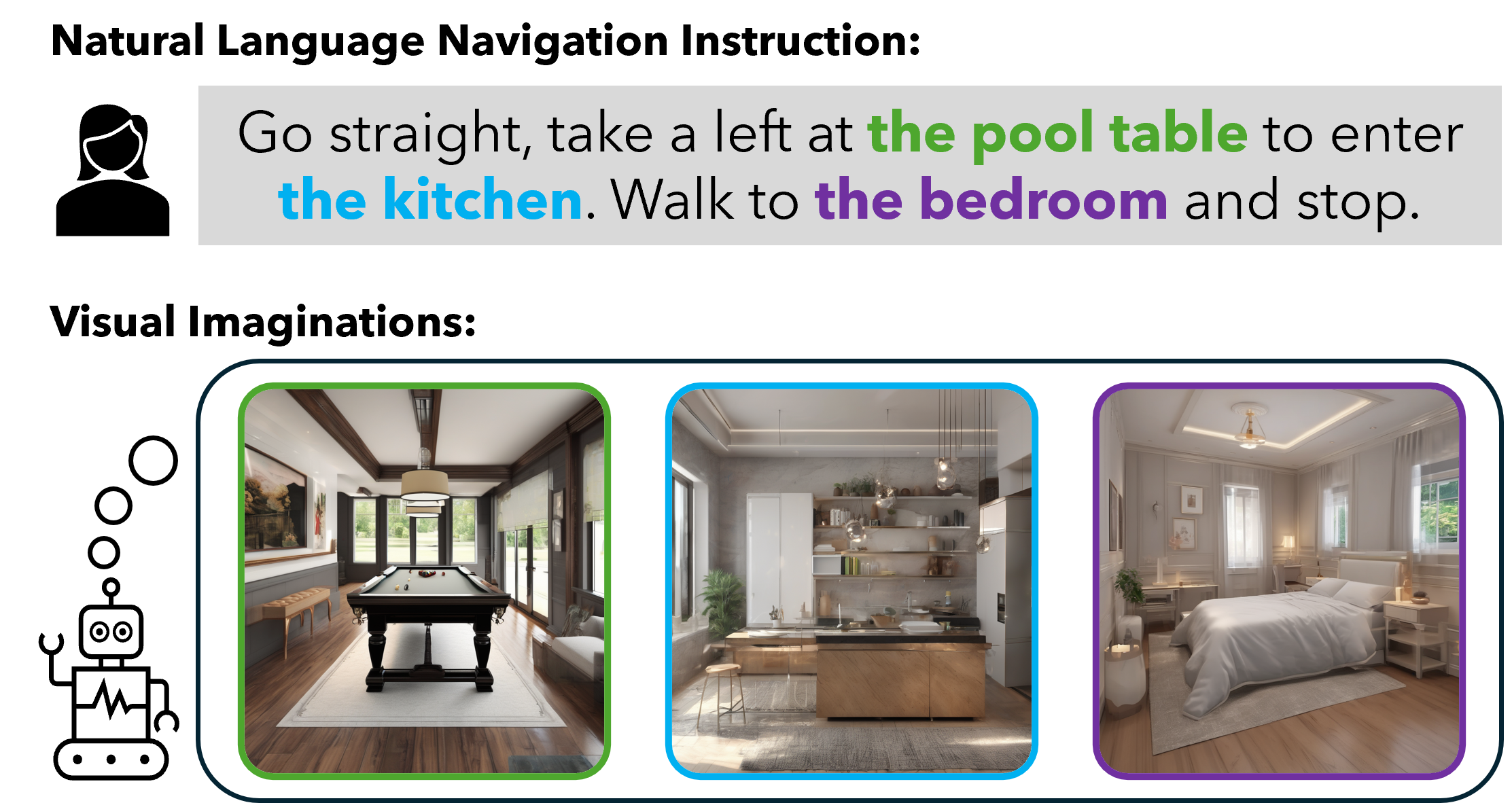}
    \caption{Illustration of visual imaginations. (Top) A natural language instruction specifying sub-goals pool table, kitchen, and bedroom. (Bottom) Visual imaginations of landmarks pool table, kitchen and bedroom referenced by the sub-goals in the instruction. In our work, we study if these visual imaginations generated using text-to-image models can improve performance in VLN.}
    \label{fig:idea_illustration}
    \vspace{-11pt}
\end{figure}

In this work, we examine whether providing visual imagery corresponding to described landmarks improves the performance of agents following natural language navigation instructions. 
Consider the natural language navigation instruction presented in Figure \ref{fig:idea_illustration} which asks an agent to ``\emph{Go straight, take a left at the pool table to enter the kitchen. Walk to the bedroom and stop}''. This instruction provides unconditional action directives like ``\emph{go straight}'' but also frequently conditions the given directions on visual landmarks in the scene such as the pool table, kitchen, and bedroom. Standard approaches to vision-and-language navigation tasks rely on learned cross-modal alignment mechanisms to \emph{implicitly} associate these noun phrases with their visual referents during navigation. However, text-to-image generation models have improved to the point that producing imagery matching the semantics of these visual references prior to navigation is plausible. We show three such generations for our running example in Figure \ref{fig:idea_illustration}. Drawing an analogy to the substantial work in cognitive science on the impact of mental imagery on task performance, we refer to these generated images as visual \emph{imaginations} and study whether providing them in addition to corresponding language-based instructions can improve the performance of vision-and-language navigation agents.

But why might we believe visual imaginations could be beneficial? In short -- text-to-image models have broad knowledge, and learning to perform semantic matching in the image domain may be an easier task than language grounding. Associating noun phrases and their visual referents is a sub-task in vision-and-language navigation (VLN) that most works address through extensive pretraining of model components on web-scale image caption datasets -- e.g., by initializing image and language encoders with CLIP \cite{radford2021learning} models. While this provides a useful warm-start, it is unclear how well the remaining cross-modal components that drive navigation retain these capabilities after downstream VLN training. Downstream datasets may cover only a small portion of relevant objects and visual grounding remains a weakness in many existing VLN models \cite{zhu2022diagnosing,yang2023behavioral}.

Directly synthesizing visual imaginations using off-the-shelf text-to-image models offers alternative training and inference mechanisms. Synthesized images have known correspondences to noun phrases in instructions that can be leveraged to reinforce visual grounding during downstream task training through auxiliary losses. Further, matching a visual imagination with environment observations reduces to an image-to-image matching task -- offering the potential to use visual imaginations as a convenient pivot between language and vision. This may prove especially effective for out-of-distribution landmarks that have limited or no support during downstream task training. For example, novel instructions referencing ``butterfly sculptures'' or ``Pulp Fiction posters'' are unlikely to be well-grounded in the training set but produce semantically relevant visual imaginations that could then be matched to observations.

Though we do not imply any direct connections, our work is also inspired by cognitive science studies which indicate mental imagery can serve as valuable cues in addition to linguistic stimuli. Several works supporting dual-coding theory \cite{Paivio1, Paivio-1965, PaivioSmytheYuille} find that concrete imagery tends to be more effective for certain types of concept learning compared to linguistic stimuli alone. Further, findings from human and animal studies \cite{REDDAN2018994, mental-rotation, Schacter2007, rats-imagery} suggest that imagining snapshots of the environment or potential trajectories before navigating can improve decision making. 

To investigate our key question, we augment agents with visual imaginations for the popular Room-2-Room (R2R) \cite{Anderson2017VisionandLanguageNI} and REVERIE \cite{qi2020reverie} tasks. We propose a VLN agent agnostic procedure to incorporate visual imaginations as additional inputs and introduce a text-imagination alignment loss to explicitly encourage visual grounding during training. We generate visual imaginations for key noun phrases in instructions that refer to visual landmarks using an off-the-shelf text-to-image generation model \cite{podell2023sdxl}.

Applying our technique to two existing models,  HAMT \cite{chen2021hamt} and DUET \cite{Chen_2022_DUET}, we find modest improvements -- R2R validation unseen success increases by roughly 1.0 and 0.6 points respectively. This translates to an improvement of 2 points in success on the test set for our modified DUET model. On REVERIE validation unseen, we improve DUET performance by 1.3 points success and 0.82 points grounding success. Our results suggest that including explicit visual depictions of referred objects can improve performance on vision-and-language navigation tasks.

\xhdr{Contributions.} We summarize our contributions as:\\[-8pt]

\begin{compactenum}[\hspace{3pt}--]
    \item We develop a pipeline for generating visual imaginations from navigation instructions and synthesize the R2R-Imagine dataset to enable studying the impact of text-to-image models on VLN agents.\\[-8pt]
    \item We propose an agent-agnostic method to incorporate visual imaginations into existing VLN agents and show improved performance for HAMT \cite{chen2021hamt} and DUET \cite{Chen_2022_DUET} across R2R and REVERIE.\\[-8pt]
    \item We provide ablations of modeling decisions to characterize the design space for adding visual imaginations.
\end{compactenum}

\csection{Related Work}
\label{sec:related-work}

\noindent \textbf{Vision-and-Language Navigation (VLN)}. Navigating agents that can operate in novel environments using free-form natural language lends itself as personal assistants, field-robots and search-and-rescue agents. Development of VLN agents has been facilitated by photorealistic simulators such as Matterport3D \cite{MP3D} and paired language-action datasets such as Room-2-Room (R2R) \cite{Anderson2017VisionandLanguageNI} and REVERIE \cite{qi2020reverie}. R2R provides fine-grained instructions conveying low-level directives such as ``Go left, then right at the hall''. REVERIE specifies coarse-grained instructions conveying broad high-level goals such as ``Adjust the picture by the lamp in the hall". Progress in VLN has broadly ranged from architectural improvements and data scaling techniques. Architectural improvements include evolution from recurrent systems \cite{Anderson2017VisionandLanguageNI} to transformer architectures \cite{recbert_Hong_2021_CVPR, chen2021hamt, Chen_2022_DUET}. HAMT \cite{chen2021hamt} uses a hierarchical transformer architecture to encode spatial and temporal information. DUET \cite{Chen_2022_DUET} maintains a topological map and performs coarse and fine scale encoding over it. Data scaling techniques include augmenting environments \cite{envdrop_tan-etal-2019-learning, hao2020learning} using an action-to-text speaker model \cite{fried2018speaker} to generate synthetic instructions, and scaling environments \cite{kamath2023new, wang2023scalevln} with additional scenes such as HM3D \cite{ramakrishnan2021habitat} and Gibson \cite{xiazamirhe2018gibsonenv} in Habitat \cite{savva2019habitat}. Recently, zero-shot or learned integrations of pretrained large language models for VLN \cite{Zhou_Hong_Wu_navgpt_2024, zhou2025navgpt} is showing promising performance. However, running large models in-the-loop incurs high computational costs.

Closely related to our work, ADAPT \cite{ADAPT} augments VLN agents with visual observations related to an instruction. Language-observation pairs are stored from the training set and multiple pairs of matching object/location nouns from instructions are retrieved at test time using CLIP \cite{radford2021learning} scores. Fundamentally, the ADAPT image base is limited to training environments whereas our imaginations cover an open distribution. Therefore, our method can be more representative of the instruction and visualize novel objects/locations. For instance, a sub-instruction of ``walk into the kitchen with blue walls'' would retrieve images of kitchens from the training environment using ADAPT, but our imaginations would convey a kitchen with blue walls.

Another adjacent work, LAD \cite{li2023layoutLAD} leverages layout dreamer and goal dreamer modules optimized around a topological graph to navigate via coarse-grained instructions. The capability of our method contrasts with LAD in three primary ways:
1) method generality; Our method is performant across models and datasets, whereas LAD is a full architecture constructed around a particular topological graph in coarse-grained settings,  2) sequential imagination: we imagine a sequence of landmarks along a path to inform sequential decision-making, whereas LAD imagines just the goal, and 3) we perform early fusion of imaginations and language whereas LAD employs late fusion.

\xhdr{Image generation in VLN.}
Several techniques have proposed generative models to predict future visual observations or entire environments in VLN \cite{koh2021pathdreamer,li2023vln-sig,li2023panogen}. Pathdreamer \cite{koh2021pathdreamer} developed a GAN-based model to predict panoramic observations after a navigation action in VLN, conditioned on prior visual observations. Agents utilizing these predicted observations in a fixed-horizon search with a learned trajectory-instruction alignment heuristic achieve non-trivial success compared to simple baselines, but fall short of planning on actual observations and standard VLN models. Unlike our approach, Pathdreamer does not condition on instruction text or explore training agents on generated imagery. Similarly, PanoGen \cite{li2023panogen} uses a diffusion model to also generate panorama observations, but conditions on image captions while doing so as a form of visual dataset augmentation -- replacing Matterport3D observations with synthetic generations with matching semantics. In contrast to our approach, this data augmentation does not provide additional inputs to the VLN models, is not conditioned on instructions, and is not active at inference time.

Related to our work is VLN-Sig \cite{li2023vln-sig}, which adapts VLN agents to predict quantized features of future visual observations given the instruction and observation history. This supports additional pretraining tasks and an auxiliary loss during downstream VLN finetuning for predicting next-step image semantics. Unlike our approach, VLN-Sig relies on the trained VLN agent to predict future image properties and focuses more on next-step prediction. In contrast, we leverage broad vision-and-language knowledge from text-to-image models and focus generating instruction-referred visual landmarks to improve agent performance.

\xhdr{Image generation in robot control.}
Taking a wider view, several methods for controlling embodied agents have leveraged text-conditioned generative techniques to provide goal specifications in the form of images or videos. For tabletop object rearrangement tasks, many works use text-to-image models to synthesize goals or sub-goals in response to natural language commands, see \cite{blackzero,dall-e-bot,gao2023can} for a non-exhaustive sample. Others produce denser conditioning, leveraging text-to-video models to generate entire potential robot trajectories \cite{du2024learning,bharadhwaj2024gen2act}. While similar in motivation to our approach, these techniques focus on tabletop manipulation settings which introduce difficult physical challenges but significantly limit the problem to constrained spaces that remain largely observed throughout operation. As such, sub-goals are often edits to existing observations rather than generations of  yet-to-be-seen landmarks as in our approach.

\begin{figure*}[t]
  \centering
  \includegraphics[width=1\linewidth]{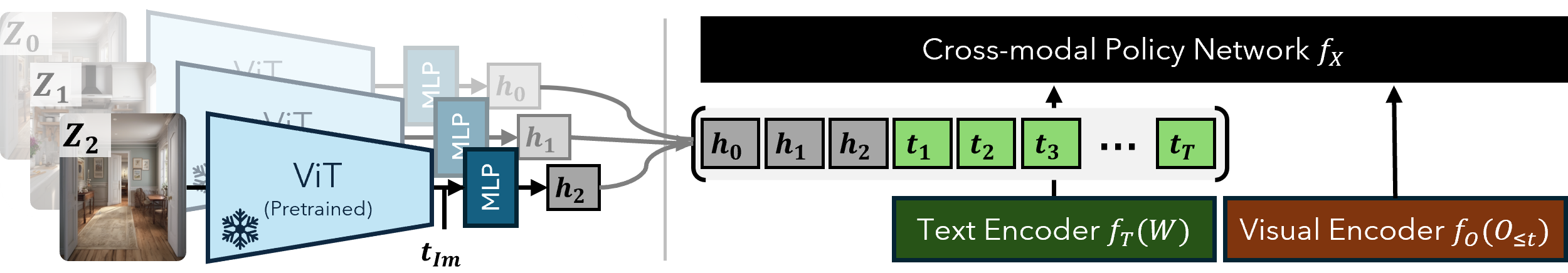}
  \caption{An overview of our approach. (Left) Imaginations generated using valid sub-instructions from an instruction as determined by our filtering scheme are first passed to a pre-trained ViT to obtain feature vectors. A type embedding $t_{Im}$ for imagination modality is then added to the features which are encoded using a 3 layer MLP to obtain imagination embeddings $h_i$. (Right) To integrate imagination modality to a VLN agent, the imagination embeddings $h_i$ are concatenated with instruction embeddings $t_i$ that are encoded using a text encoder $f_T(W)$. The concatenated imagination-text embeddings are passed to the VLN agent's cross-modal encoder $f_X$ along with visual embeddings to predict a distribution over the agent's action space.}
  \vspace{-11pt}
  \label{fig:integration_method}
\end{figure*}

\begin{figure}[t]
  \centering
    \includegraphics[width=\linewidth]{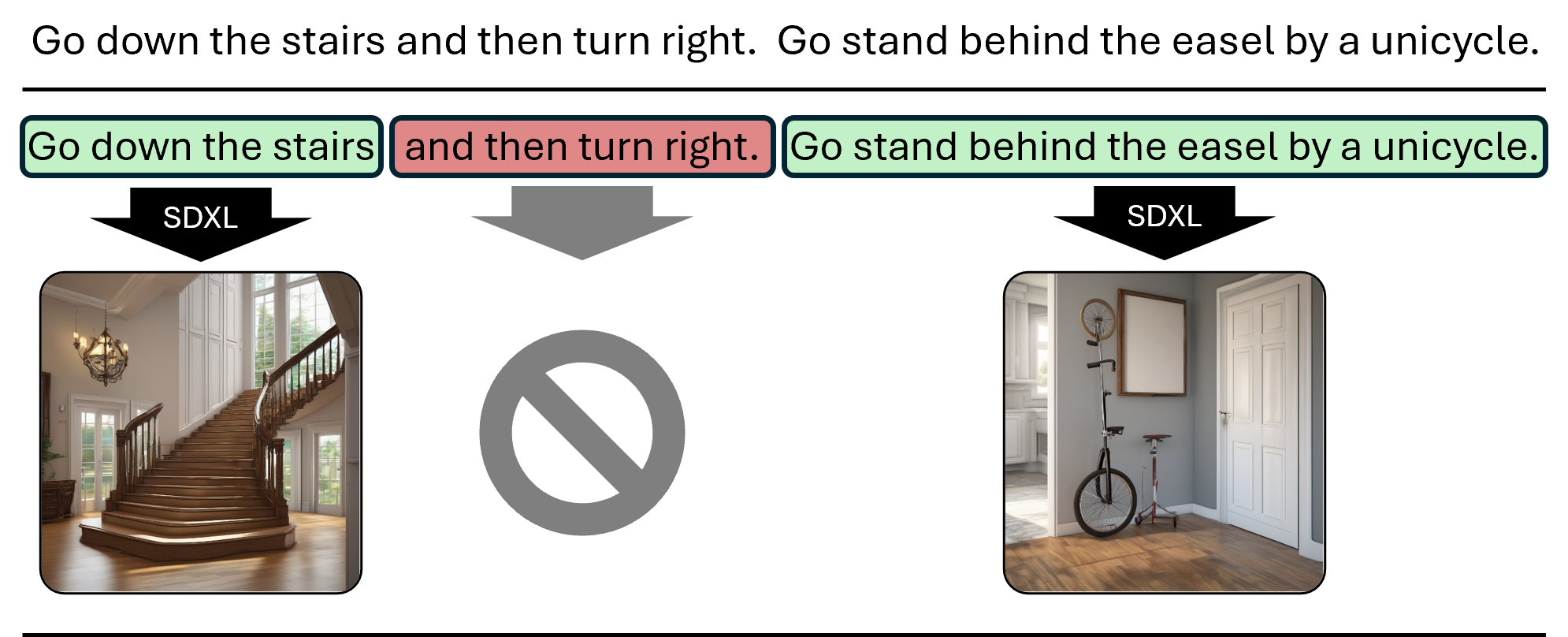}
    \caption{Example instruction segmentation, filtering, and image generation. Instructions are segmented to sub-instructions leveraging FG-R2R \cite{fgr2r} and then filtered to remove phrases referring to uninformative nouns (e.g., ``right''). We produce visual imaginations using SDXL \cite{podell2023sdxl} for remaining sub-instructions.}
  \label{fig:R2R-Imagine}
\end{figure}

\csection{Methodology}
\label{sec:method}

Our proposed methodology consists of two components -- (1) a visual imagination generation pipeline and (2) a model-agnostic approach to integrating visual imaginations into existing Vision-and-Language Navigation (VLN) agents. Before defining these, we start with establishing relevant notation for the VLN task.

\xhdr{Problem definition}. In the standard VLN task \cite{Anderson2017VisionandLanguageNI}, an agent is prompted with a natural language navigation instruction and must make a sequence of actions to navigate along the described trajectory based on visual observations. We denote the $L$-word instruction as $W = (w_1, w_2, \cdots, w_{L})$ and the visual observation at time step $t$ as a panoramic observation $O_t$ consisting of $K$ single view images ${O}_t = (I^t_1, \cdots, I^t_{K})$. Following prior work, we consider $K=36$ corresponding to images from 12 headings and 3 pitches. Amongst the  single view images, a subset $\mathcal{A}_t\subset O_t$ correspond to navigable directions. Given the instruction $W$ and history of observations $O_{\leq t}$ until time $t$, the agent selects an action $a_t \in \mathcal{A}_t \cup \{\texttt{stop}\}$ and accordingly either moves or terminates the episode. Generally, the VLN agent is parameterized as a policy network $\pi_\theta(a_t \mid W, O_{\leq t})$ with parameters $\theta$ learned via general pretraining and a combination of imitation learning and reinforcement learning \cite{chen2021hamt, Chen_2022_DUET}.

\csubsection{Generating Visual Imaginations}
\label{subsec:R2R-imagine}

To generate visual imaginations of potential landmarks, we consider the structure of navigation instructions. Typically, these consist of a sequence of sub-instructions describing intermediate navigation steps, which may or may not refer to visual landmarks. For instance, a sub-instruction ``go past the couch'' implies a visual landmark while ``go straight then left'' does not. We use FG-R2R \cite{fgr2r} to segment instructions, resulting in an average of 3.66 sub-instructions per instruction for the R2R\cite{Anderson2017VisionandLanguageNI} training set. For a given instruction $W$ decomposed into $m$ segments, we denote these as sub-instructions $\mathcal{S} = (S_0, \cdots, S_m)$.

\xhdr{Sub-instruction filtering scheme.} As sub-instructions may not always contain references to visual landmarks, we filter sub-instructions prior to image generation. First, we ignore any sub-instruction that lacks a noun phrase using Spacy \cite{spacy} and then noun phrases are further filtered by a manual blacklist to avoid uninformative landmark generation. For example, we exclude detected noun phrases rooted on non-visual terms like counts (`one'), directions (`left'), or ambiguous object pronouns (``it'') as these lack specificity to produce meaningful visual landmarks. After filtering, we denote the remaining sub-instructions $\mathcal{S}' \subset \mathcal{S}$.

\xhdr{Visual imagination generation.} The filtered sub-instructions are passed through a text conditioned diffusion model as illustrated in Fig. \ref{fig:R2R-Imagine}. We use SDXL \cite{podell2023sdxl} as the diffusion model with positive and negative prompts chosen to steer generation towards indoor real-estate environments akin to typical environments in Matterport3D \cite{MP3D}. For example, providing ``indoor'' and ``real estate'' as positives prompts and ``humans'' and ``collage'' as negatives. See the supplementary for a full list.
For a given instruction, we denote the generated images for valid sub-instructions as $\mathcal{Z}=\{~Z_i~|~S_i \in \mathcal{S}'~\}$. We colloquially refer to these generated images as imaginations and generate them for the entire R2R dataset. The resulting R2R-Imagine dataset consists of over 41k synthesized imaginations. This amounts to an average of 2.96 imaginations per instruction.

\csubsection{Integration with Existing Models}
\label{subsec:alg}

We consider augmentation with visual imagination to be a model agnostic approach and integrate it in two prior models --  HAMT \cite{chen2021hamt} and DUET \cite{Chen_2022_DUET}. Our integration approach consists of an imagination encoder, an alignment-promoting auxiliary loss, and a finetuning regimen (Fig. \ref{fig:integration_method}). 
Without loss of generality, we consider a decomposition of any VLN agent into an instruction encoder $f_T(W)$, an observation/history encoder $f_O(O_{\leq t})$, and a cross-modal policy network $f_X( f_T(W),  f_O(O_{\leq t}))$ that combines the encoder outputs to predict a distribution over actions. Note that these may be fairly complex mechanisms and include explicit history encoding modules separate from encoding the current observation; however, this abstraction is sufficient here.

\xhdr{Imagination encoder and integration.} Given an instruction $W$ with corresponding imaginations $\mathcal{Z}$, we independently encode each imagination $Z_i$ using a pretrained vision transformer (ViT) \cite{dosovitskiy2020vit} to produce a $d$-dim embedding vector. These are further transformed by a three-layer MLP such that the $i$th imagination embedding can be written as

\begin{equation}
  h_i = \mbox{MLP}(~\mbox{ViT}(Z_i) + t_{Im}~ )
  \label{eq:imag_enc}
\end{equation}

\noindent where $t_{Im}$ is the type embedding for imagination modality. As we consider visual imaginations to be proxies for sub-instructions, the full set of imagination embeddings $\mathcal{H}$ are concatenated to the text encodings such that our general VLN agent structure now computes $f_X(~ [f_T(W), \mathcal{H}],  f_O(O_{\leq t})~)$ where $[\cdot, \cdot]$ denotes a concatenation. Taking the DUET model as an example, the imagination embeddings are concatenated to the instruction encoding prior to being passed to the coarse-scale and fine-scale cross-modal encoders.

\xhdr{Auxiliary alignment loss.} 
We add an auxiliary loss to align the representations of imaginations and their corresponding sub-instruction's noun phrases. For a given sub-instruction $S_i$, we denote the mean embedding vector for noun phrase tokens in $S_i$ from the text encoder $f_T$ as $\bar{S}_i$. To align representations, we compute a cosine similarity loss between the corresponding imagination $Z_i$ encoded as $h_i$ and $\bar{S}_i$. This is averaged over all $N_{Im}$ imagination-subinstruction pairs $(Z_i, S_i)$ in a batch,

\begin{equation}
 \mathcal{L}_{cos} = \frac{1}{N_{Im}} \sum_{(Z_i, S_i)\in B} \left( 1 - \frac{h_i\cdot \bar{S}_i}{\lVert h_i \rVert \lVert \bar{S}_i \rVert}\right).
 \label{eq:cos_sim}
\end{equation}

\noindent During finetuning, we combine this auxiliary loss $\mathcal{L}_{\text{cos}}$ with the losses from the base agent with a scaling factor $\lambda$ such that the overall loss is  $\mathcal{L}_{\text{base}} + \lambda \mathcal{L}_{\text{cos}}$.

\begin{table*}
  \centering
  \small
  \tabcolsep=0.08cm
  \caption{Comparison of our approach with selected prior work on the R2R dataset. Methods that use additional visual data beyond MP3D \cite{MP3D} are annotated with $\dag$ and are not directly comparable with other approaches. Note that BEVBert* additionally uses panoramic depth images. Adding our visual imagination approach to HAMT and DUET base models (gray rows) leads to improved success rate (SR) and success weighted by inverse path length (SPL) metrics on the val-unseen split. For DUET, this effect also persists on the test set. We highlight these improvements over base models using \textbf{bold} numbers.}
  \label{tab:r2r_vln_agent_comparison}
  \resizebox{\textwidth}{!}{
  \begin{tabular}{l p{1em} cccc  p{1em} cccc p{1em} cccc p{0.25em}} \toprule
    \multirow{2}{*}{Methods} && \multicolumn{4}{c}{Validation Seen} && \multicolumn{4}{c}{Validation Unseen} && \multicolumn{4}{c}{Test Unseen} \\\cmidrule(lr){3-6}\cmidrule(lr){8-11}\cmidrule(lr){13-16}
    && TL & NE$\downarrow$ & SR$\uparrow$ & SPL$\uparrow$ && TL & NE$\downarrow$ & SR$\uparrow$ & SPL$\uparrow$ && TL & NE$\downarrow$ & SR$\uparrow$ & SPL$\uparrow$ \\ \midrule
    PREVALENT \cite{hao2020learning} && 10.32 & 3.67 & 69 & 65 && 10.19 & 4.71 & 58 & 53 && 10.51 & 5.30 & 54 & 51 \\
    RecBERT \cite{Hong_2021_CVPR} && 11.13 & 2.90 & 72 & 68 && 12.01 & 3.93 & 63 & 57 && 12.35 & 4.09 & 63 & 57 \\ 
    ADAPT \cite{ADAPT} && 10.97 & 2.54 & 76 & 72 && 12.21 & 3.77 & 64 & 58 && 12.99 & 3.79 & 65 & 59 \\
    BEVBert* \cite{an2023bevbert} && 13.56 & 2.17 & 81 & 74 && 14.55 & 2.81 & 75 & 64 && 15.87 & 3.13 & 73 & 62 \\
          MARVAL \cite{kamath2023new}$^\dag$
&& 10.60 & 2.99 & 73 & 69
&& 10.15 & 4.06 & 65 & 61
&& 10.22 & 4.18 & 62 & 58 \\
      ScaleVLN \cite{wang2023scalevln}$^\dag$ &&
11.90 & {2.16} & {87} & {80}
&& 12.40 & {2.34} & {87} & {79} &&  14.27 & {2.73} & {83} & {77} \\
  \midrule
    HAMT \cite{chen2021hamt}  && \textbf{11.15} & 2.51 & 75.61 & 72.18 && \textbf{11.46} & 3.62 & 66.24 & 61.51 && \textbf{12.27} & 3.93 & 65 & 60 \\ 
     \rowcolor{gray!15} HAMT-Imagine (ours) && 11.30 & \textbf{2.36} & \textbf{77.16} & \textbf{73.72} && 11.81 & \textbf{3.58} & \textbf{67.26} & \textbf{62.02} && 12.66 & \textbf{3.89} & 65 & 60 \\ 
 \midrule
            DUET \cite{Chen_2022_DUET} && \textbf{12.33} & 2.28 & 78.84 & 72.88 && \textbf{13.94} & 3.31 & 71.52 & 60.41 && \textbf{14.73} & 3.65 & 69 & 59 \\
      \rowcolor{gray!15} DUET-Imagine (ours)  && 12.93 & \textbf{2.19} & \textbf{79.9} & \textbf{73.75} && 14.35 & \textbf{3.19} & \textbf{72.12} & \textbf{60.48} && 15.35 & \textbf{3.52} & \textbf{71} & \textbf{60} \\

\bottomrule
  \end{tabular}}
\end{table*}

\xhdr{Finetuning regimen.}
We start with fully-trained checkpoints of prior models as our initial weights. Depending on the base method, these parameters may be the result of multi-stage pretraining and finetuning regimens -- e.g., leveraging synthetic PREVALENT \cite{hao2020learning} data as prescribed in \cite{chen2021hamt}. The imagination encoder is integrated to the base model and finetuned with the R2R-Imagine dataset. To mitigate catastrophic forgetting, we perform our training in three stages. First, we train only the newly introduced imagination encoder $MLP$ along with type embeddings $t^{Im}$ with the remaining parameters frozen. Then, all the modules are trained jointly with a decreased learning rate for the base model. Finally, all the parameters are trained at a common learning rate. See supplementary materials for details on learning rates and schedules.

\csection{Experiments}
\label{sec:experiments}

Using the R2R dataset, we evaluate the impact of visual imaginations on our agent's navigation ability, perform ablation studies to explore the design space, and illustrate qualitative impacts of visual imaginations. Using the REVERIE dataset, we study the generalizability of our technique to a fine-grained instruction setting.

\csubsection{Experimental setup}
\label{subsec:exp-setup}

\noindent \textbf{R2R dataset}. R2R \cite{Anderson2017VisionandLanguageNI} is constructed using Matterport3D \cite{MP3D} simulator. It consists of 90 indoor environments with 10,567 panoramas (each panorama consists of 36 single-view images) which act as nodes in a navigation graph. There are 7,189 trajectories each with 3 instructions and a corresponding ground truth trajectory. The dataset is split into train (4675 trajectories), val-seen (340 trajectories), val-unseen (783 trajectories), and test (1391 trajectories). The train and val-seen split share environments while val-unseen and test have different environments.

\xhdr{R2R-Imagine dataset}. The curation process of R2R-Imagine is prescribed in section \ref{subsec:R2R-imagine}. The number of imagination images are 41,558 for train, 3,055 for val-seen, 6,857 for val-unseen, and 12,412 for test. Our imaginations have a resolution of 1024x1024 and a single imagination generation on a single H100 GPU takes 3.2 seconds on average. We intend to release the dataset upon acceptance.

\xhdr{REVERIE dataset}. REVERIE \cite{qi2020reverie} is constructed using Matterport3D and consists of high-level goals instead of finer grained step-by-step instructions. These instructions generally convey the target object and its location. The agent has to rely more on exploration relative to R2R. To be successful, agents must reach and identify the referent object. To synthesize imaginations for REVERIE, we treat the entire instruction as a single sub-instruction and follow the same recipe used for R2R.

\xhdr{Evaluation metrics}. We employ standard VLN metrics \cite{Ilharco2019GeneralEF}. Success rate (SR) is the ratio of instruction-trajectories where the agent stops within a 3 meter radius from the goal and success normalized by inverse path length (SPL) helps distinguish successful episodes with shorter path lengths from longer path lengths. Navigation error (NE) is the distance between an agent's final position and the goal in meters. Trajectory length (TL) is the length of an agent's traversed path in meters. REVERIE \cite{qi2020reverie} additionally defines grounding metrics Remote Grounding Success (RGS) and RGS penalized by path length (RGSPL).

\xhdr{HAMT and DUET base models.} We apply our approach to two baseline representative architectures -- HAMT \cite{chen2021hamt} and DUET \cite{Chen_2022_DUET}. HAMT is hierarchical and transformer-based; instructions, observations, and trajectory history are encoded separately using unimodal transformers and then combined in a single cross-modal transformer to predict actions. DUET incrementally forms a topographic map of the environment over time and uses dual fine- and coarse-grained cross-modal attention mechanisms to associate language embeddings with the local observation and global topographic features. The results are then fused to predict actions. For both base models, we inject visual imaginations via concatenation with the text embeddings (Sec.~\ref{subsec:alg}). We use provided checkpoints for pretrained models.

\xhdr{Implementation details}. We follow base agent \cite{chen2021hamt, Chen_2022_DUET} settings unless specified otherwise. We use a standard off-the-shelf pretrained ViT-B/16 \cite{dosovitskiy2020vit} to encode imaginations for both HAMT and DUET. Our VLN-Imagine agents are finetuned for 100k iterations in three stages as noted in section \ref{subsec:alg}. The finetuning process is carried out using a Tesla V100 GPU with a batch size of 8 and takes $\sim$ 1.5 days for each agent. For combining the proposed auxiliary  loss function with the base model losses, we empirically set $\lambda{=}0.5$.

\csubsection{Results}
\label{subsec:results}

Our experiments demonstrate that providing generated images of instruction landmarks (\textit{imaginations}) improves the performance of vision-and-language navigation models (Tab. \ref{tab:r2r_vln_agent_comparison}). We highlight our key findings below.

\xhdr{VLN agents perform better with imaginations.}
VLN agents equipped with imaginations perform better than VLN agents without by 0.6-1.0 SR and up to 0.5 SPL on R2R val-unseen (Tab. \ref{tab:r2r_vln_agent_comparison}). This improvement coincides with lower navigation error (NE) across both models and all dataset splits. We contextualize these results relative to agent architectures that are representative of the VLN literature. We additionally report a distinct family of techniques that scale training environments to boost performance by reducing overfitting -- MARVAL \cite{kamath2023new} and ScaleVLN \cite{wang2023scalevln}. We consider approaches of this class orthogonal to our study and instead use visual data only from Matterport3D \cite{MP3D}.

\begin{table}
	\centering
    \tabcolsep=0.10cm
	\caption{Performance comparison of our approach on REVERIE with baseline. We observe improvements of our method across all metrics on DUET when provided coarse-grained instructions.}
	\label{tab:reverie_performance}
	\begin{tabular}{p{3.5cm} cccc} \toprule
		& SR$\uparrow$ & SPL$\uparrow$ & RGS$\uparrow$ & RGSPL$\uparrow$ \\\midrule
            DUET & 46.98 & 33.73 & 32.15 & 23.03 \\ 
  		DUET-Imagine (ours) & 48.28 & 33.76 & 32.97 & 23.25 \\ 
            \bottomrule
	\end{tabular}
\end{table}

\begin{table}
	\centering
	\tabcolsep=0.30cm
	\caption{Role of imaginations; HAMT-Imagine. Null imaginations refers to testing with no imaginations, wrong imaginations refers to testing with imaginations sampled from different instructions. We observe aligned imaginations are important to navigation performance. In addition, results with no imaginations are better than baseline implying a regularization effect during training.}
	\label{tab:validate_imagination}
	\begin{tabular}{p{4.5cm} cc} \toprule		
		& SR$\uparrow$ & SPL$\uparrow$ \\ \midrule
  		Baseline (HAMT) & 66.24 & 61.51 \\ 
            Null imaginations & 66.92 & 61.89 \\
            Wrong imaginations & 66.24 & 61.02 \\ 
  		Correct imaginations (ours) & 67.26 & 62.02 \\ \bottomrule
	\end{tabular}
\end{table}
\begin{table}
	\centering
	\tabcolsep=0.30cm
	\caption{Sequential \vs goal imagination; HAMT-Imagine. We study the effect of providing just the imagination belonging to the final valid sub-instruction relative to full valid imaginations. We observe providing just the goal imagination improves performance significantly compared to baseline but shows lower performance relative to providing full imaginations.}
	\label{tab:final_goal_imagination}
	\begin{tabular}{p{4.5cm} cc} \toprule
		& SR$\uparrow$ & SPL$\uparrow$ \\\midrule
  		Baseline (HAMT) & 66.24 & 61.51 \\ 
            Goal imagination & 66.79 & 61.58 \\ 
            Full imaginations (ours) & 67.26 & 62.02 \\ \bottomrule
	\end{tabular}
\end{table}

\xhdr{Imaginations are useful in coarse-grained instruction navigation.} 
 Our method is performant when natural language instructions convey high-level objectives (REVERIE, Tab. \ref{tab:reverie_performance}); navigation performance improves by 1.3 SR and grounding performance improves by 0.82 RGS. We hypothesize that imaginations not only aid in reaching the goal but also in identifying and grounding the target concept amongst others objects in the scene. 

\xhdr{VLN agents benefit from both imagination \textit{training} and \textit{inference}.}
When nullifying imaginations at test-time, VLN-Imagine agents still surpass baseline performance, but to a lesser extent (Tab. \ref{tab:validate_imagination}). We implement this nullification by setting the imagination attention masks to zero in the cross-modal encoder. We hypothesize that imagination-based training provides a regularizing effect on the overall VLN agent during fine-tuning, revealing a research avenue for strong and performant test-time model inference.
We further find that imaginations must be aligned with the instruction to provide maximal benefit; when fed with imaginations randomly-sampled from the R2R-Imagine dataset, VLN-Imagine agents perform worse than baseline (Tab. \ref{tab:validate_imagination}).

\xhdr{Sequential imagination outperforms goal imagination.} 
Sequential refers to imaginations generated over multiple sub-instructions (our method). Goal imagination refers to imagining just the final non-filtered sub-instruction. We show these results in Tab. \ref{tab:final_goal_imagination} -- while goal imagination outperforms the baseline by 0.5 SR, sequential imagination outperforms the baseline by 1.0 SR. Goal imagination may be particularly useful in guiding the agent to stop at the destination. While distinct from our setting, such goal imagination has parallels to ImageNav \cite{krantz2022instancespecific}, a visual navigation task that conveys goals solely using exemplar images.

\xhdr{General-purpose vision encoders are sufficient imagination encoders.}
We compare encoding imaginations with off-the-shelf ViT~\cite{dosovitskiy2020vit} \vs ViT fine-tuned on R2R navigation episodes \cite{chen2021hamt}. We find comparable performance (Tab. \ref{tab:validate_vit}) and choose to employ off-the-shelf ViT in our final model. This is motivated to maximize generality and to avoid potential issues relating to catastrophic forgetting. For example, generalization of frozen ViT is demonstrated in Zeng et al. \cite{zengpoliformer}.

\xhdr{Aligning imaginations to instructions benefits VLN-imagine agents.}
Aligning the imagination embedding to the instruction embedding as an auxiliary loss leads to a 0.5 SR and a 0.4 SPL increase over no auxiliary loss (Tab. \ref{tab:loss_comparison}). We also consider formulating this loss as contrastive -- do negative samples enable a stronger alignment and thus better downstream performance? We find that an InfoNCE loss \cite{oord2018representation} yields no significant difference relative to our cosine similarity loss (Eq. \ref{eq:cos_sim}). We construct negative samples using mean noun phrase embeddings from imaginations of other instructions in the batch. The overall loss is  $\mathcal{L}_{\text{base}} + \lambda \mathcal{L}_{\text{InfoNCE}}$ where we set $\lambda$ to 0.2 empirically.

\begin{table}
	\centering
	\tabcolsep=0.20cm
	\caption{Comparison between ViT models. We train both HAMT-Imagine and DUET-Imagine agents with a ViT finetuned in-the-loop with HAMT and an off-the-shelf pretrained ViT. We observe no consistent effects across the both agents.}
	\label{tab:validate_vit}
	\begin{tabular}{p{2.5cm} cccc} \toprule
		 & \multicolumn{2}{c}{ViT fine-tuned} & \multicolumn{2}{c}{ViT off-the-shelf} \\
		& SR$\uparrow$ & SPL$\uparrow$ & SR$\uparrow$ & SPL$\uparrow$ \\ \midrule
  		HAMT-Imagine & 67.31 & 61.5 & 67.26 & 62.02 \\ 
            DUET-Imagine & 72.58 & 60.29 & 72.12 & 60.48 \\ 
            \bottomrule
	\end{tabular}
\end{table}
\begin{table}
	\centering
	\tabcolsep=0.30cm
	\caption{Loss ablation. We study the effect of negative samples in our alignment loss by comparing our approach of using cosine loss with InfoNCE \cite{oord2018representation} loss. In addition, we train the agent with no auxiliary loss represented by ``No loss''. We do not observe a significant difference between contrastive and cosine losses.}
	\label{tab:loss_comparison}
	\begin{tabular}{p{4.5cm} cc} \toprule
		& SR$\uparrow$ & SPL$\uparrow$ \\\midrule
            No loss & 66.75 & 61.60 \\ 
            $\mathcal{L}_{\mbox{\scriptsize \text{InfoNCE}}}$ & 67.18 & 61.84 \\ 
  		$\mathcal{L}_{\text{cos}}$ (ours) & 67.26 & 62.02 \\ 
            \bottomrule
	\end{tabular}
\end{table}
\begin{table}
	\centering
	\tabcolsep=0.10cm
	\caption{Open vocabulary object detection accuracy (\%) of imaginations to noun phrases from sub-instructions. Our imaginations represent the noun phrases with a high accuracy. In val-seen, in 98.43\% of sub-instructions, at least 1 noun phrase is detected and in 94.51\% of sub-instructions, all noun phrases are detected.}
	\label{tab:imagination_fidelity}
	\begin{tabular}{l c cc} \toprule
		\multirow{2}{*}{} & \multirow{2}{*}{\text{Sub-instructions}} & \multicolumn{2}{c}{\text{Noun Phrase Detection (\%)}} \\ 
        \cmidrule(lr){3-4}
		& & \(\geq\)1 detected & All detected \\\midrule
  		\text{Val-seen} & 2985 & 98.43 & 94.51 \\ 
        \text{Val-unseen} & 6700 & 98.99 & 95.34 \\ 
        \text{Test} & 12075 & 98.93 & 95.12 \\ \bottomrule
	\end{tabular}
\end{table}

\xhdr{Imaginations correctly represent noun phrases in a sub-instruction.} To evaluate fidelity of our imaginations, we run an open-vocabulary object detector (LangSAM \cite{langsam}) on our imaginations to test if the noun phrases contained in the corresponding sub-instruction from R2R are detected successfully. We see in Tab.~\ref{tab:imagination_fidelity} that at least one noun phrase is detected in our imaginations for 98.78\% of sub-instructions and all noun phrases are detected for 94.99\% of sub-instructions. We conclude our imaginations provide a high coverage of relevant concepts across splits.

\csubsection{Qualitative visualizations}

In this section, we provide a qualitative investigation of how imaginations may be used as a pivot between language tokens and visual observations from the environment by examining the attention patterns in HAMT-Imagine. 

For a given instruction, we randomly select an imagination / sub-instruction pair and examine the agent's trajectory to identify the first occurrence of the reference object. We then examine the attention distributions between language tokens, the visual imagination, and the panoramic observation. As there are many attention heads across multiple layers in the cross-modal encoder, we structure our study by focusing on those attention heads where sub-instruction tokens are attended to most strongly by their corresponding visual imagination and then examine how the visual imagination attends to observations. 

\begin{figure*}[t]
  \centering
  \includegraphics[width=\linewidth]{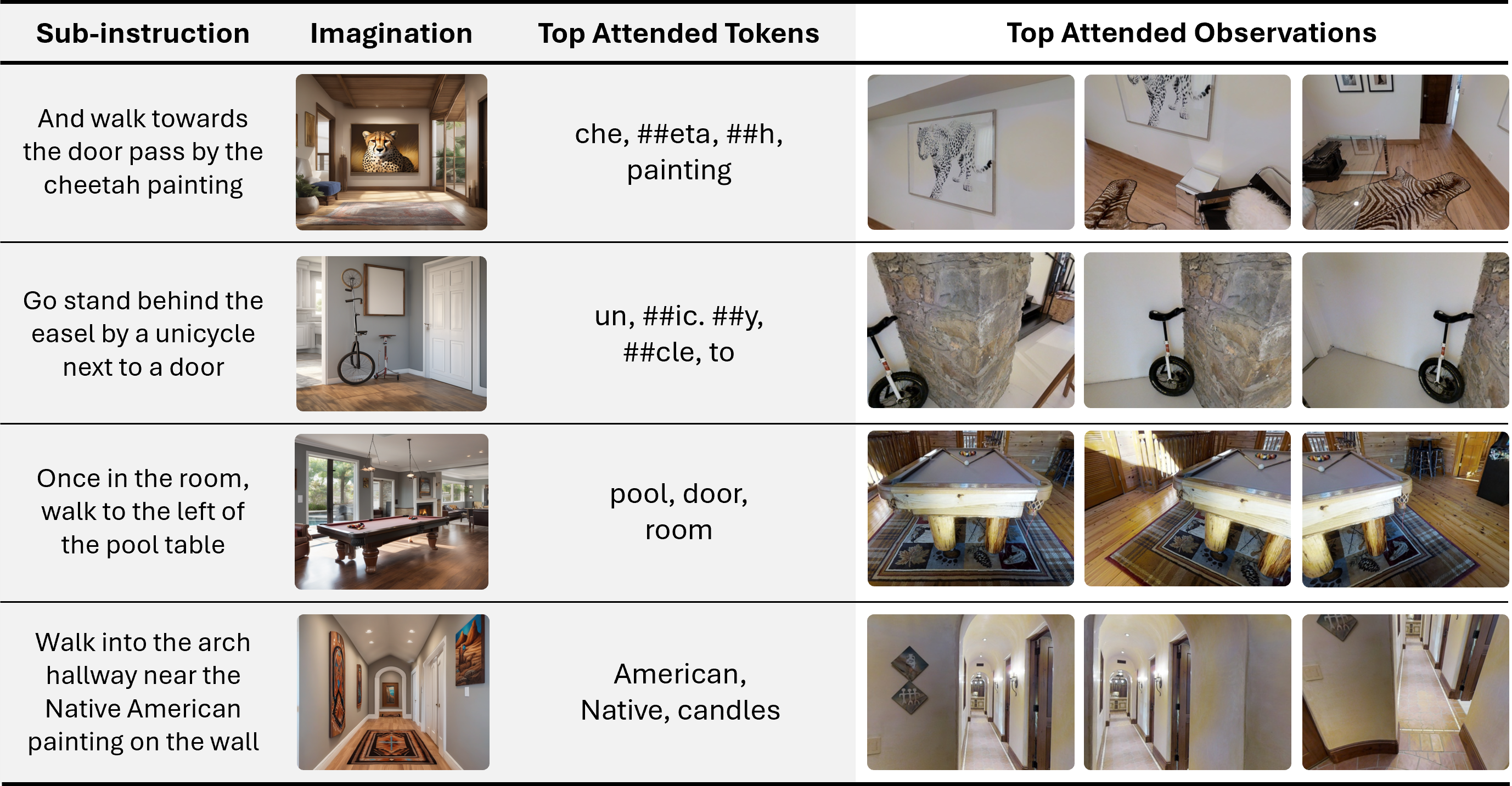} 
    \caption{Qualitative examples showing imaginations as pivots between language and observation images. The first column contains sub-instruction from a random instruction from R2R, the second column contains the imagination generated using the sub-instruction. The third and fourth columns show highest attended language tokens and observation images from an attention head in HAMT's cross-modal transformer at a time step the associated observations are first visible. In the second example (row 2), the sub-instruction references ``unicycle'' which is captured in the imagination along with neighboring nouns ``easel'' and ``door''. We observe that in a head where top attending language tokens to the imagination query are references to nouns associated with the sub-instruction, its top attended observations to the imagination query are images of the same concept (``unicycle''). In this example, the imagination of a unicycle is being used to associate language tokens belonging to ``unicycle'' to observations of unicycle hinting at the utility of imaginations in navigation.}
  \label{fig:qualitative_visualization}
\end{figure*}

We then visualize sub-instruction and imagination pairs along with their top attended language tokens and visual observations in Fig.~\ref{fig:qualitative_visualization}. By construction, these are time steps where the referent is visible and the visual imagination is attending to the appropriate noun phrase. As such, our examples primarily demonstrate how the imagination learns to associate with the visual observations. We focus on this aspect as our auxiliary loss already explicitly encourages alignment between visual imaginations and  sub-instruction.

From the examples in Fig.~\ref{fig:qualitative_visualization}, we make two remarks. \textit{Imaginations attend most strongly to observations of matching semantic concepts}. In row 1, the imagination and the retrieved observations all depict a cheetah (either a painting or floor mat). \textit{Imagination attention scores mirror the proper disambiguation of concepts}. In row 2, the sub-instruction (and thus the imagination) refer to a unicycle. While both a bicycle and unicycle are observed during execution, attention scores are maximized by the observations containing the unicycle.
These patterns do not directly imply a causal relation between the attention scores and eventual decision making of the VLN agent. Rather, they illustrate how imaginations can serve as a pivot between language and observations and afford disambiguation of related concepts.

\csection{Conclusion}
\label{sec:conclusion}

In this work, we show that visual representations of instructions, or visual imaginations, play an additive role in training superior navigating agents for vision-and-language navigation (VLN) tasks. We generate a sequence of visual imaginations depicting sub-instruction landmarks using a text-to-image diffusion model. Our approach is model-agnostic and we integrate these imaginations into VLN agents using an imagination encoder, an auxiliary loss, and a fine-tuning training step. We apply this approach to two representative VLN agents across two datasets and observe performance improvements of around 1 SR and 0.5 SPL.

\xhdr{Limitations.} Generating and encoding imaginations increases the computational cost of running VLN agents. This is particularly relevant when running agents on-device for real-world robotic deployments. However, we note that this process is performed upfront, unlike techniques that depend on step-wise observations like novel view synthesis (NVS). Stepping more broadly, by design, our imaginations are not grounded in the environment. Personalized reasoning, such as the unique naming of objects and locations, cannot be directly imagined in our framework. Life-long learning of persistent groundings is an open question not just in generating imaginations, but in the VLN space at large.

\xhdr{Future directions.} A natural question that arises from our research is the exact complementary nature of the roles played by visual imaginations and language instructions. This is a rich area for future work. Similar studies relating language and visual observations have been conducted in \cite{VLN-nonsense} and \cite{zhu2022diagnosing}. Our paper demonstrated the fundamental utility of imagination in VLN agents. Beyond this lies interesting extensions, such as how imagination can help bridge the simulation-to-reality (Sim2Real) gap for VLN agents and whether imaginations can unlock the performance of VLN world models through image-image reasoning.

\xhdr{Acknowledgments.} We thank Zijiao Yang and Robert Yelle for their assistance with simulator issues and technical support for high performance computing infrastructure.


{
    \small
    \bibliographystyle{ieeenat_fullname}
    \bibliography{main}
}

\clearpage
\maketitlesupplementary

\section{Model details}
\label{sec:supp_implementation}

\subsection{Integration with base agents}

\begin{figure}[t]
  \centering
    \includegraphics[width=\linewidth]{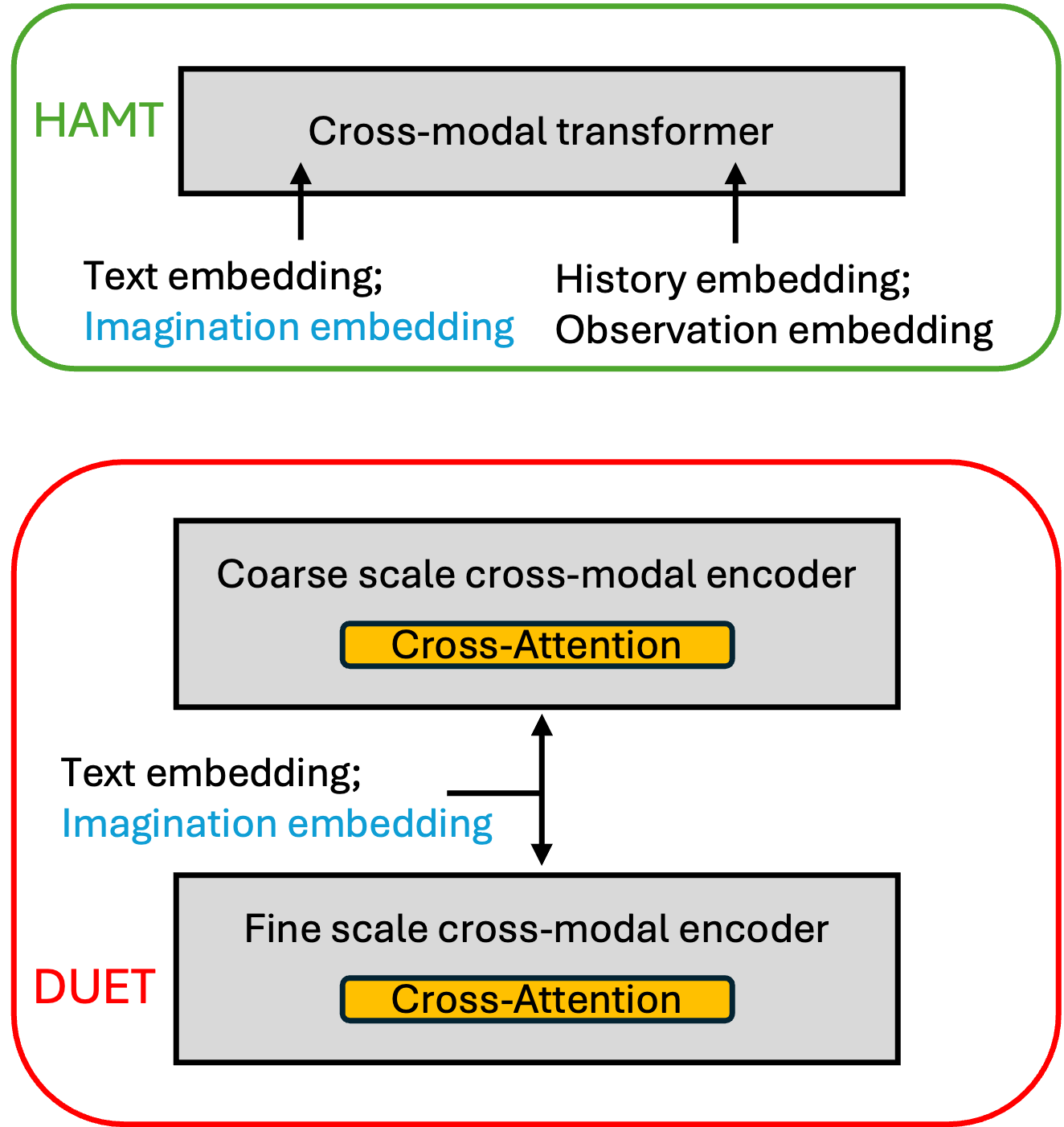}
    \caption{Integration of imagination embeddings to HAMT and DUET. The imagination embeddings are concatenated with language embeddings before passing through cross-modal encoders.}
  \label{fig:supp_integration}
\end{figure}

 We show our integration mechanism in detail for HAMT and DUET in Fig. \ref{fig:supp_integration}. In HAMT, the imagination embedding vector $h$ is concatenated with language embeddings and passed to the language modality branch of HAMT's CMT. The visual modality branch with observations and history along with the action prediction network is retained as is. In DUET, the imagination embedding vector $h$ is concatenated with language embeddings and passed to the coarse-scale cross-modal encoder to perform global action prediction, and to the fine-scale cross-modal encoder to perform local action prediction. The fusion between the dual-scale action predictions along with rest of the architecture is retained as is.

\subsection{Training routine}

In order to mitigate catastrophic forgetting, we train our agent for 100k iterations in three stages:
\begin{compactitem}
    \item First, we train only the imagination encoder $MLP$ along with type embeddings $t^{Im}$ at a learning rate of $1e-4$ for 25\% of iterations. Rest of the modules from the base agent are frozen.
    \item Then, all the modules are trained jointly for the next 25\% of iterations. The imagination encoder parameters are trained at a learning rate of $5e-5$ and the base agent parameters are trained at $1e-6$. 
    \item Finally, all the parameters are trained at a common learning rate of $1e-6$ for rest of the training.
\end{compactitem}

\subsection{MLP architecture}

Our $MLP$ consists of three fully connected (FC) layers with ReLU activations after the first two layers. Our input dimension is 768, hidden dimension is 512 and output dimension is 768. We apply a dropout layer with rate 0.15 to the input and omit bias terms in all layers.

\section{Imagination guidance prompts}
\label{sec:supp_guidance_prompts}

We use positive prompts to guide the generations towards indoor environments and negative prompts to exclude concepts such as humans and outdoor environments. The complete prompts are listed below:
\begin{compactitem}
    \item Positive prompts: indoor, house, realistic, real estate.
    \item Negative prompts: outdoor, text, humans, man, woman, boy, girl, collage.
\end{compactitem}

\section{Additional experiments}
\label{sec:supp_additional_exp}

\begin{table}
	\centering
	\tabcolsep=0.20cm
    \renewcommand*{\arraystretch}{1.25}
	\caption{Visual \vs textual representations; Training with mean sub-instruction embeddings in place of imagination embeddings leads to a performance drop albeit better than baseline implying the advantages of visual imaginations.}
	\label{tab:visual_vs_text_rep}
	\begin{tabular}{p{4.5cm} cc} \toprule
		& SR$\uparrow$ & SPL$\uparrow$ \\\midrule
  		Baseline (HAMT) & 66.24 & 61.51 \\ 
            HAMT-Imagine (ours) & 67.26 & 62.02 \\ 
            Sub-instrs only & 66.67 & 61.50 \\ \bottomrule
	\end{tabular}
    \vspace{5.5pt}
\end{table}
\xhdr{Visual representations of landmarks outperform textual representations.} We study the effect of replacing visual imaginations with textual sub-instructions in this experiment. To do this, we train a HAMT based agent similar to HAMT-Imagine but in place of imagination embeddings, we use mean sub-instruction embeddings. In Tab. \ref{tab:visual_vs_text_rep}, we notice textual representations while leading to better navigation performance than baseline is still inferior to visual imaginations by 0.59 SR and 0.52 SPL. This might imply imaginations play a complementary role to language in our setting.

\begin{table}
	\centering
	\tabcolsep=0.60cm
    \renewcommand*{\arraystretch}{1.25}
	\caption{Design ablations for imagination encoder. We observe best performance when imaginations are encoded by an MLP and concatenated along with instruction encodings.}
	\label{tab:design_ablation}
            \begin{tabular}{p{2.5cm} cc} \toprule
                Design & SR$\uparrow$ & SPL$\uparrow$ \\ \midrule
                HAMT-Imagine & 67.26 & 62.02 \\ \midrule
                Transformer & 66.75 & 61.64 \\ 
                Visual concat & 66.54 & 61.38\\ \bottomrule
            \end{tabular}
\end{table}

\begin{table}
	\centering
	\tabcolsep=0.60cm
    \renewcommand*{\arraystretch}{1.25}
	\caption{Early fusion \vs late fusion. Our early fusion leads to better navigation success.}
	\label{tab:lad_late_fusion}
            \begin{tabular}{p{2.5cm} cc} \toprule
                Design & SR$\uparrow$ & SPL$\uparrow$ \\ \midrule
                DUET-Imagine (early fusion) & 72.12 & 60.48 \\ \midrule
                LAD late fusion & 71.73 & 60.44 \\ \bottomrule
            \end{tabular}
\end{table}
\xhdr{Simple MLP is a sufficient imagination encoder when concatenated with language.} We experiment with alternate designs for encoding visual imaginations in Tab. \ref{tab:design_ablation}. We use a transformer (row 2) with positional encodings to encode imaginations before passing them to the cross-modal encoder in HAMT. MLP as imagination encoders leads to a better performance (rows 1, 2) in our setting. One possible explanation is that having fewer parameters can help reduce overfitting, which VLN agents are susceptible to due to limited data availability. Additionally, imaginations when concatenated with instructions as opposed to visual observations improves effectiveness of our agent (rows 1, 3). We hypothesize that imaginations act as ``visual instructions'' such that concatenating them with instruction might aid in relevant inductive biases. Finally, we compare our early fusion approach of integrating imagination features with that of LAD's \cite{li2023layoutLAD} late fusion using our DUET-imagine agent. LAD fuses imagination features with DUET's global features using a node-specific learned weight to compute an action distribution. We incorporate their approach to our DUET-imagine agent and contrast it with our early fusion. We report in Tab. \ref{tab:lad_late_fusion}, early fusion provides an improvement of 0.39 SR and 0.04 SPL.

\section{Qualitative visualizations}
\label{sec:supp_qualitative}

\begin{figure*}[t]
  \centering
  \includegraphics[width=\linewidth]{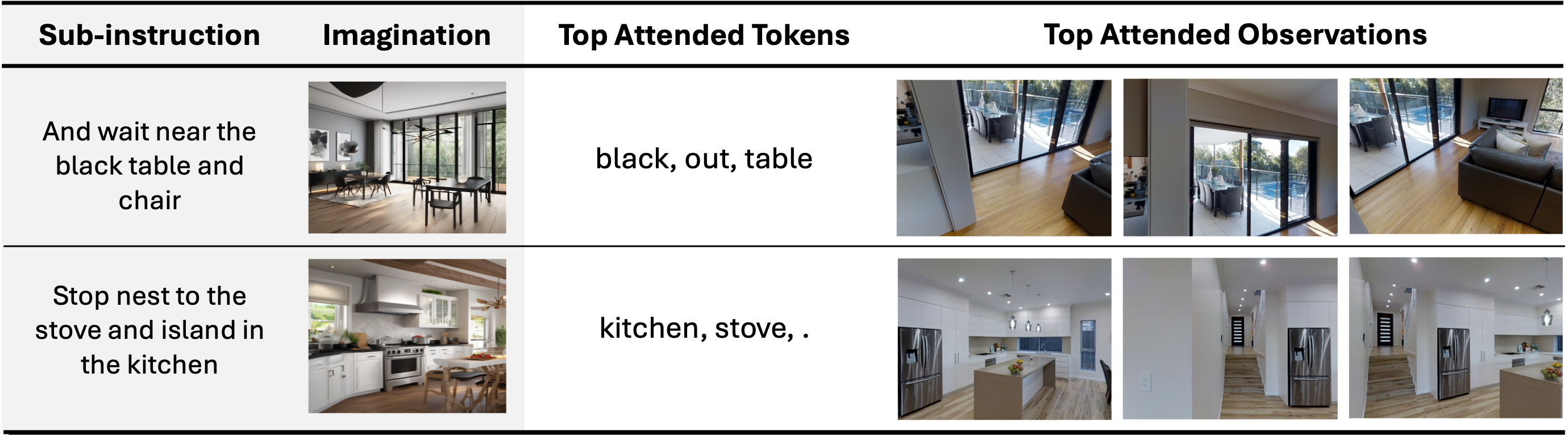}
    \caption{Additional qualitative examples illustrating the role of imaginations as pivots between language and observation images. The first example (row 1) illustrates the potential application of imaginations in disambiguating destinations with similar looking objects (black colored objects). The second example (row 2) showcases the potential ability of imaginations to act as a pivot between language and observations of kitchen and stove.}
  \label{fig:supp_qualitative_visualization}
\end{figure*}

First, we show additional qualitative visualizations of top attended concepts in Fig. \ref{fig:supp_qualitative_visualization}. In the first example (row 1), the imagination captures noun phrases ``black table'' and ``chair''. The imagination strongly attends to related language tokens ``black'', ``table'' and a noun phrase from a different sub-instruction ``out''. The top attended observation images by the selected attention head contains black table and chairs. As can be partially seen in the observation images, the neighborhood contains other black objects that can mislead an agent hinting at the potential of imaginations in disambiguating similar concepts. In the second example, the imagination captures a stove and kitchen. The top attended language tokens by a selected attention head are kitchen and stove. Its corresponding top attended observation tokens capture the same concepts as well.

\begin{figure*}[t]
  \centering
  \includegraphics[width=\linewidth]{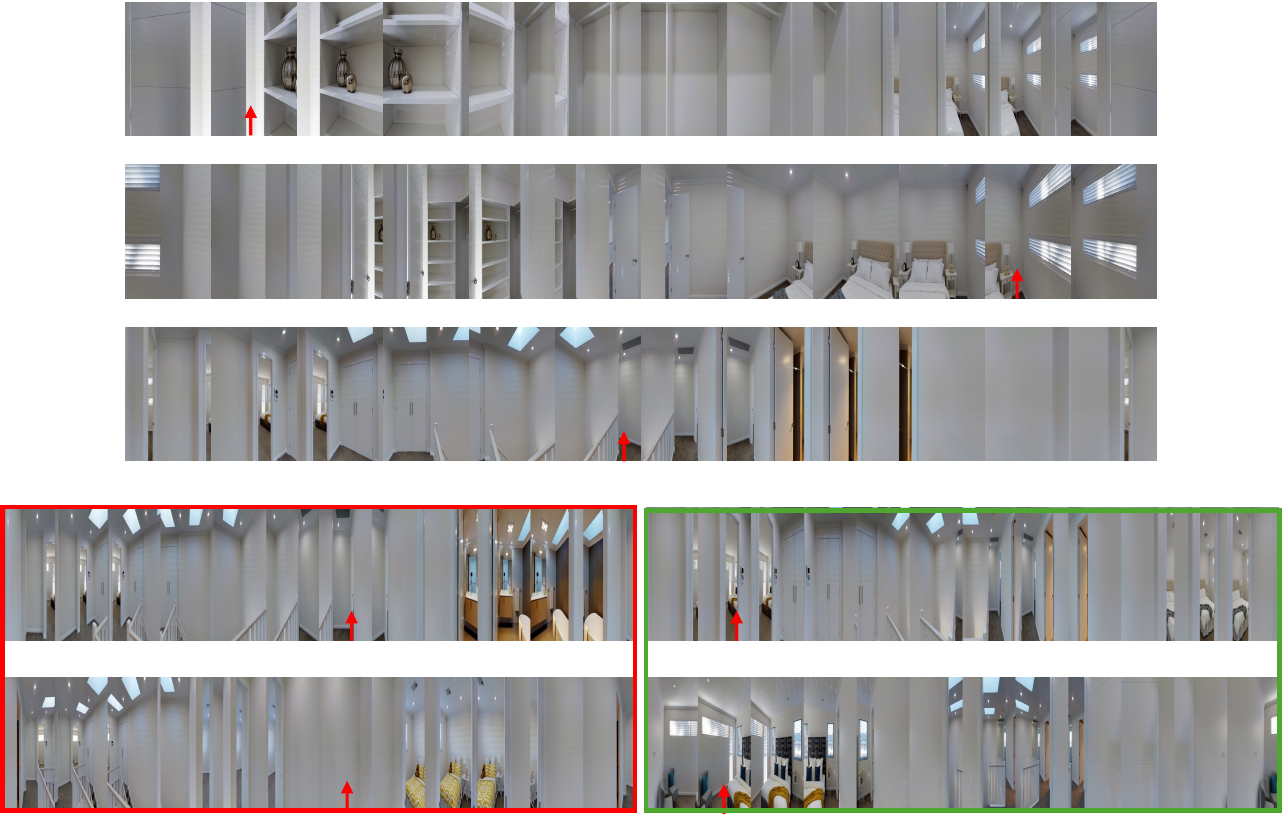}
    \caption{Qualitative visualization of trajectory from HAMT and HAMT-Imagine for sample 431\_1. In the first three timesteps, both the agents are aligned. However, HAMT (left) is unable to decide between the two bedrooms in close vicinity whereas our agent HAMT-Imagine (right) is able to disambiguate between similar concepts to arrive at the correct bedroom with a golden blanket.}
  \label{fig:supp_qualitative_traj}
\end{figure*}

We also illustrate a sample trajectory of our HAMT-Imagine agent in comparison with the baseline HAMT agent in Fig. \ref{fig:supp_qualitative_traj}. We consider sample 431\_1 from R2R val-unseen with instruction ``Walk into the bedroom area. Walk passed the bed and through the door. Walk down the hallway and into the bedroom with the striped bed backboard and golden blanket laying on top." The baseline HAMT agent stops prematurely adjacent to a different bedroom along the path. Our agent, which is provided a synthesized imagination containing a striped backboard and golden blanket, successfully continues past the incorrect bedroom and stops at the correct bedroom. We hypothesize our agent is able to use the imagination to disambiguate between similar concepts in this example.

\end{document}